\documentclass[english]{article}
\usepackage{amsfonts}
\usepackage{amsmath}
\usepackage{amssymb}
\usepackage{amsthm}
\usepackage{verbatim}
\usepackage{subfig}
\usepackage[colorlinks,citecolor=blue,urlcolor=blue,linkcolor=red]{hyperref}
\usepackage{fullpage}
\usepackage[affil-it]{authblk}

\usepackage{graphicx}

\DeclareFontFamily{U}{mathx}{\hyphenchar\font45}
\DeclareFontShape{U}{mathx}{m}{n}{<-> mathx10}{}
\DeclareSymbolFont{mathx}{U}{mathx}{m}{n}
\DeclareMathAccent{\widebar}{0}{mathx}{"73}

\renewcommand{\phi}{\varphi}
\renewcommand{\epsilon}{\varepsilon}

\newcommand{\DS}{\mathrm{DS}}
\newcommand{\DC}{\mathrm{DC}}
\newcommand{\MS}{\mathrm{MS}}
\newcommand{\Sp}{\mathrm{Sp}}
\newcommand{\Sn}{\mathrm{Sn}}

\begin{document}
\title{A Deep Learning Approach to Digitally Stain
Optical Coherence Tomography Images of the Optic Nerve Head 
}

\author[1]{Sripad Krishna Devalla}
\author[2]{Jean-Martial Mari}
\author[4,5]{Tin A. Tun}
\author[4,6,7]{Nicholas G. Strouthidis}
\author[4]{Tin Aung}
\author[3 $\star$]{Alexandre H. Thiery}
\author[1,4 $\star$]{Michael J. A. Girard}

\affil[1]{Ophthalmic Engineering \& Innovation Laboratory, Department of Biomedical Engineering, Faculty of Engineering, National University of Singapore, Singapore.
}
\affil[2]{GePaSud, Universit{\'e} de la Polyn{\'e}sie francaise, Tahiti, French Polynesia.}
\affil[3]{Department of Statistics and Applied Probability, National University of Singapore, Singapore.}
\affil[4]{Singapore Eye Research Institute, Singapore National Eye Centre, Singapore.}
\affil[5]{Yong Loo Lin School of Medicine, National University of Singapore, Singapore.}
\affil[6]{NIHR Biomedical Research Centre at Moorfields Eye Hospital NHS Foundation Trust and UCL Institute of Ophthalmology, London, United Kingdom}
\affil[7]{Discipline of Clinical Ophthalmology and Eye Health, University of Sydney, Sydney, New South Wales, Australia}
\bigskip
\affil[$\star$]{Both authors contributed equally and are both corresponding authors.}

\maketitle


%
%
\begin{abstract}
\noindent
{\bf Purpose.} To develop a deep learning approach to digitally-stain optical coherence tomography (OCT) images of the optic nerve head (ONH).\\

\noindent
{\bf Methods.} A horizontal B-scan was acquired through the center of the ONH using OCT (Spectralis) for 1 eye of each of 100 subjects (40 normal \& 60 glaucoma). All images were enhanced using adaptive compensation. A custom deep learning network was then designed and trained with the compensated images to digitally stain (i.e. highlight) 6 tissue layers of the ONH. The accuracy of our algorithm was assessed (against manual segmentations) using the Dice coefficient, sensitivity, and specificity. We further studied how compensation and the number of training images affected the performance of our algorithm.\\

\noindent  
{\bf Results.} For images it had not yet assessed, our algorithm was able to digitally stain the retinal nerve fiber layer + prelamina, the retinal pigment epithelium, all other retinal layers, the choroid, and the peripapillary sclera and lamina cribrosa. For all tissues, the mean dice coefficient was $0.84 \pm 0.03$, the mean sensitivity $0.92 \pm 0.03$, and the mean specificity $0.99 \pm0.00$. Our algorithm performed significantly better when compensated images were used for training ($p < 0.001$). Increasing the number of images (from 10 to 40) to train our algorithm did not significantly improve performance ($p < 0.05$), except for the RPE.\\

\noindent
{\bf Conclusion.} Our deep learning algorithm can simultaneously stain neural and connective tissues in ONH images. Our approach offers a framework to automatically measure multiple key structural parameters of the ONH that may be critical to improve glaucoma management.

\end{abstract}

%
%
\section{Introduction}
\label{sec.intro}
In glaucoma, the optic nerve head (ONH) exhibits complex structural changes including, but not limited to, thinning of the retinal nerve fiber layer (RNFL) \cite{RN1}; changes in choroidal thickness  \cite{RN2,RN3} minimum-rim-width \cite{RN4}, and lamina cribrosa depth \cite{RN5}; and scleral canal expansion and bowing \cite{RN6,RN7}. If all these structural parameters (and their changes) could be measured automatically with optical coherence tomography (OCT), it could considerably assist clinicians in their day-to-day management of glaucoma.\\

For OCT research, manual segmentation has remained the gold-standard to extract structural information of the ONH, and this is especially true for deeper connective tissues \cite{RN8,RN9}. However, manual segmentation is time consuming, prone to bias, and unsuitable in a clinical setting \cite{RN10,RN11}. While several techniques have been proposed to automatically segment some (but not all) ONH tissues in OCT images \cite{RN10}--\cite{RN20}, each tissue currently requires its own processing algorithm. This lack of ``universal" approach may limit the clinical translation and appeal for these algorithms.\\

Furthermore, the quality of automated segmentations/delineations largely depends on that of the OCT images. Poor deep-tissue visibility and shadow artifacts \cite{RN21} in OCT images as a result of light attenuation makes the development of robust segmentation tools difficult. With the advent of swept-source OCT \cite{RN22}, enhanced depth imaging \cite{RN23,RN24,RN25} and compensation technology \cite{RN26} the quality of OCT images has been improved, opening the door to new possibilities. Recently, our group has developed a post-processing technique that, when combined with compensation, could digitally-stain (highlight) neural and connective tissues in OCT images of the ONH. However, this approach remains limited, as it cannot identify each ONH tissue separately, and in some cases, requires manual inputs \cite{RN27}.\\

In this study, we aimed to develop a custom deep learning algorithm to automatically and simultaneously stain six important neural and connective tissue structures in OCT images of the ONH. We hope to offer a framework to automatically extract key structural information that has remained difficult to obtain in OCT scans of the ONH.

\section{Methods}
\subsection{Patient Recruitment}
A total of 100 subjects were recruited at the Singapore National Eye Centre. All subjects gave written informed consent, and the study adhered to the tenets of the Declaration of Helsinki and was approved by the institutional board of the hospital. The subject population consisted of 40 normal (healthy) controls, 41 subjects with primary open angle glaucoma (POAG), and 19 primary angle closure glaucoma (PACG). Inclusion criteria for healthy controls were: intraocular pressure (IOP)  21 mmHg, healthy optic nerves with vertical cup-disc ratio (CDR) less than or equal to 0.5 and normal visual fields. POAG was defined as glaucomatous optic neuropathy (GON; characterized as loss of neuroretinal rim with vertical CDR $> .7$ and/or focal notching with nerve fiber layer defect attributable to glaucoma and/or asymmetry of CDR between eyes $> 0.2$) with repeatable glaucomatous visual field defects. PACG was defined as the presence of GON with compatible visual field loss, in association with a closed anterior chamber angle and/or peripheral anterior synechiae in at least one eye. A closed anterior chamber angle was defined as the posterior trabecular meshwork not being visible in at least 180 of anterior chamber angle.

\subsection{Optical Coherence Tomography Imaging}
OCT imaging was performed on seated subjects under dark room conditions after dilation with tropicamide $1\%$ solution. Images were acquired by a single operator (TAT). The diagnosis was masked with the right ONH being imaged in all the subjects, unless the inclusion criteria were met only in the left eye, in which case the left eye was imaged. A horizontal B-scan (0) of $8.9$ mm (composed of 768 A-scans) was acquired through the center of the ONH for all the subjects using spectral-domain OCT (Spectralis, Heidelberg Engineering, Heidelberg, Germany). Data averaging was set to 48 and enhanced depth imaging (EDI) was used for all scans. 

\subsection{Correction of Light Attenuation Using Adaptive Compensation}
In order to remove the effects of light attenuation from OCT images, all B-scans were post-processed using adaptive compensation (AC) \cite{RN26}. For OCT images of the ONH, AC has been shown to e.g. remove blood vessel shadows, improve tissue contrast, and increase the visibility of several features of the ONH \cite{RN28,RN29}. For all B-scans, we used a threshold exponent of 12 (to limit noise over-amplification at high depth), and a contrast exponent of 2 (to improve overall image contrast) \cite{RN23}.

\subsection{Manual Segmentation of OCT Images}
\begin{figure}[h]
    \centering
    \includegraphics[width=1.\textwidth]{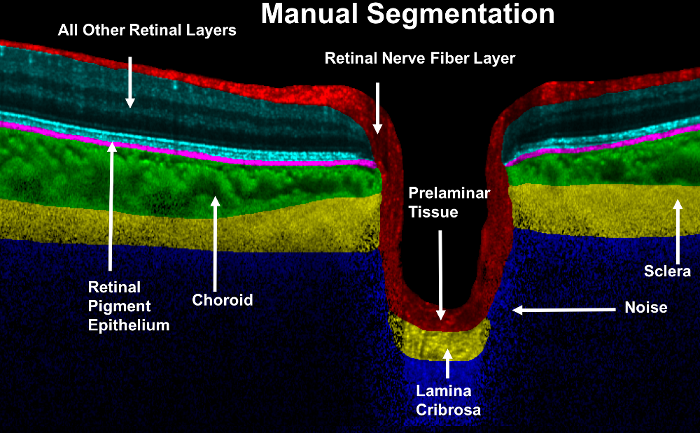}
    \caption{Manual segmentation of a compensated OCT image of the ONH. The RNFL and prelamina are shown in red, the RPE in pink, all other retinal layers in cyan, the choroid in green, the peripapillary sclera and LC in yellow, and noise in blue. }
    \label{fig:1}
\end{figure}

We performed manual segmentation of all compensated OCT images in order to: 1) train our digital staining algorithm to identify and highlight tissues; and to 2) validate the accuracy of our approach. Specifically, each compensated OCT image was manually segmented by an expert observer (SD) using Amira (version 5.4, FEI, Hillsboro, OR) to identify the following classes: (1) the RNFL and the prelamina (in red, Figure \ref{fig:1}); (2) the retinal pigment epithelium (RPE; in purple); (3) all other retinal layers (in cyan); (4) the choroid (in green); and (5) the peripapillary sclera and the LC (in yellow). Noise (below the peripapillary sclera and LC) was color-coded in blue. Note that in most cases, a full-thickness segmentation of the peripapillary sclera and of the LC was not possible due to limited visibility \cite{RN29}. Therefore, we only segmented the visible portions of the sclera/LC as detected from the compensated OCT signal, and no effort was made to capture their accurate thickness. The manual segmentation assigned a label (defined between 1 and 6) to each pixel of each OCT image to indicate the tissue class.

\subsection{Digital Staining of the ONH using Deep Learning}
We developed a custom deep learning approach to automatically stain tissues in OCT images of the ONH. Specifically, we used a 2D convolutional neural network (CNN) -- a technique that has not yet been applied to OCT images of the ONH,  but that has already been proven successful in segmenting tissues from other medical imaging modalities \cite{RN30,RN31}. Briefly, our 2D CNN was trained with manually segmented OCT images to recognize the most representative features of each tissue (present in small image patches). When our CNN was presented with an OCT image it had not yet seen, it was able to identify patches in the new image with features that ``matched" those from the training set; each patch was then assigned a color code in its center (or probability to belong to a given tissue; 1 color per tissue between 1 and 6), to generate a digitally stained image of the ONH.

\subsection{Network Architecture}
\begin{figure}[h]
    \centering
    \includegraphics[width=1.\textwidth]{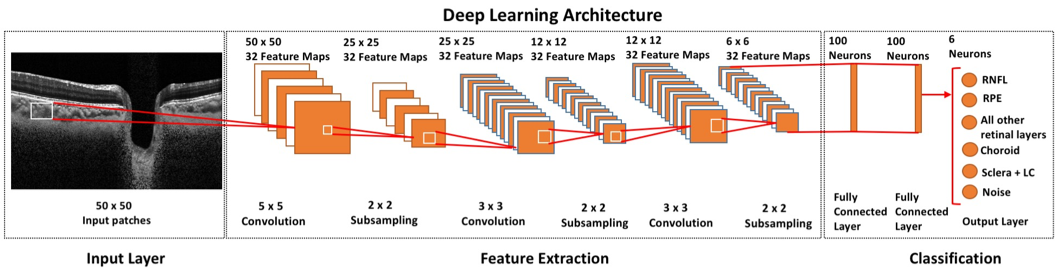}
    \caption{The deep learning architecture is an 8-layer convolution neural network (CNN) composed of three convolution layers, three max-pooling (sub-sampling) layers and two fully connected layers}
    \label{fig:2}
\end{figure}

For this study, we used a 8-layer convolutional neural network (CNN) that was composed of 3 convolution layers, 3 max-pooling layers and 2 fully connected layers (see detailed architecture in Figure \ref{fig:2}). Overlapping patches (size: 50 x 50 pixels; stride length: 1 pixel) were extracted from each OCT image (size: 497 x 768 pixels) and fed as a single channel gray-scale image to the input layer.  Each of the 3 convolution layers extracted 32 feature maps with filters of size 5 x 5, 3 x 3, and 3 x 3 pixels, respectively. The feature maps from the hidden layers (all layers except the input and output layers) were activated using a rectified linear unit (ReLu) function. Two fully connected layers with 100 neurons each were used to connect all the activations in the previous layers and funnel their excitations to the output layer. The output layer had 6 neurons, each corresponding to one class of tissue (i.e., RNFL, RPE, all other retinal layers, choroid, peripapillary sclera + LC, and noise). A softmax activation function was then applied to the output layer to obtain the class-wise probabilities (to belong to a given class) for each patch. For simplicity, every patch was assigned the class label with the highest probability in its centre.\\

The proposed CNN comprised of $130 000$ neurons and was able to learn the initially unknown weights and biases during training using a standard cross-entropy loss function and an ADAM gradient descent optimization algorithm (learning rate: 0.001) \cite{RN32}. To reduce overfitting, a dropout of $35\%$ was used. The loss function was scaled during the training using class weights for each output class of tissue to circumvent the fact that tissues covering large areas in OCT images (e.g. RNFL + prelamina) were represented by more patches. Specifically, the class weights assigned to each class of tissue were inversely proportional to the number of patches representing it in the training set (i.e. the more patches representing a particular class of tissue, the lesser its weight). Due to the limited size of our dataset, we performed online data augmentation (as is common in machine learning) by rotating (10 degrees clockwise and counter-clockwise), flipping (horizontally) and translating (5 pixels; vertically and horizontally) our patches. The proposed CNN learnt the specific features for each class of tissue in batches of 50 patches over 100 epochs.

\subsection{Training and Testing of our CNN}
The available dataset of 100 B-scans (40 healthy, 60 glaucoma) was split into two sets: (1) the training set that was used to train our network to recognize individual ONH tissues (composed of equal number of compensated healthy and glaucoma OCT images with their corresponding manual segmentations); and (2) the testing set (composed of the remaining number of compensated healthy and glaucoma OCT images with their corresponding manual segmentation) that was used to evaluate the performance of the digital staining algorithm with respect to manual segmentation.\\

In each of the training sets, only $80\%$ of the images were used to train our network. The remaining $20\%$ (referred to as the ``validation set") was used to adjust and fine-tune the model hyper-parameters in order to provide the best digital staining accuracy. This fine-tuned model was then used to evaluate its performance using the corresponding testing set.\\

Four training set sizes, of 10, 20, 30, and 40 B-scans, respectively, were used to test their effects on digital staining accuracy. Furthermore, in order to test the repeatability of our approach, five training sets were defined for each training set size, and 5 tests were performed with the remaining images. From the available 100 images, it was possible to obtain five distinct training sets of size 10 and 20. However, the training sets of size 30 and 40 had some images repeated across sets.\\

Finally, to study the effect of compensation on the performance of our digital staining algorithm, we repeated the entire process during which baseline (uncompensated) images were used instead for training.

\subsection{ Digital Staining Performance: Qualitative Assessment}
All digitally stained images were reviewed manually by an expert observer (SD) for all training sets (with compensated and uncompensated images) and compared (qualitatively) with their corresponding manual segmentations.

\subsection{Digital Staining Performance: Quantitative Assessment}

In order to estimate the accuracy of digital staining in identifying individual ONH tissues in OCT images during the testing process, 3 metrics were used for each tissue including: (1) the Dice Coefficient ($\DC$); (2) Sensitivity ($\Sn$); and (3) Specificity ($\Sp$). It is important to emphasize that these metrics could not be directly applied to the peripapillary sclera and LC as their through-thickness visibility varied considerably across images. Instead, staining of the sclera and LC was assessed qualitatively.\\

The dice coefficient is a standard measure of similarity between two shapes and was used to assess the ``overlap" between manual segmentation and digital staining. The dice coefficient is typically defined between 0 and 1, where 1 represents a perfect overlap and 0 no overlap. The dice coefficient  was calculated for each tissue $i$ (1: RNFL and prelamina, 2: RPE, 3: all other retinal layers, and 4: choroid), and for each B-scan in each testing set. It was defined as
\begin{align} \label{eq.DC}
\DC_i =  2 \, \times \, \frac{\left| \DS_i \cap \MS_i \right|}{ \left| \DS_i \right| + \left| \MS_i\right|}
\end{align}
where $\DS_i$ is the set of pixels representing the ONH tissue $i$ in the digitally stained B-scan, and $\MS_i$ is that in the corresponding manually segmented B-scan. Specificity can assess the false predictions made by digital staining, and was calculated for each ONH tissue $i$ and for each B-scan in each testing set as
\begin{align} \label{eq.Sp}
\Sp_i =  \frac{\left| \overline{\DS}_i \cap \overline{\MS}_i \right|}{\left| \overline{\MS}_i\right|}
\end{align}
where $\overline{\DS}_i$ is the set of all pixels not belonging to tissue $i$ in the digitally stained B-scan and $\overline{\MS}_i$ is that in the corresponding manually segmented B-scan. Sensitivity can assess the ability of digital staining to accurately stain a given ONH tissue, and was calculated for each ONH tissue $i$ and for each B-scan in each testing set. It was defined as:
\begin{align} \label{eq.Sn}
\Sn_i =  \frac{\left| \DS_i \cap \MS_i \right|}{\left| \MS_i\right|}
\end{align}
Both specificity and sensitivity were reported on a scale of 0--1. 

\subsection{Effect of Training Set Size on Digital Staining Accuracy}
We used a one-way analysis of variance (ANOVA) to assess differences in dice coefficients, sensitivities and specificities (mean) for a given tissue across training set sizes (data were pooled for a given training set size). Statistical significance was set at $p < 0.05$. 

\subsection{Effect of Compensation on Digital Staining Accuracy}
We used paired t-tests to assess whether our digital staining algorithm exhibited improved performance when trained with compensated images (as opposed to uncompensated images). Specifically, for a training set size of 10, we used t-tests to assess the differences in the mean values of dice coefficients, sensitivities, and specificities for each tissue (data were pooled for the training set of size 10).

\section{Results}
\subsection{Qualitative Analysis}
\begin{figure}[h]
    \centering
    \includegraphics[width=1.\textwidth]{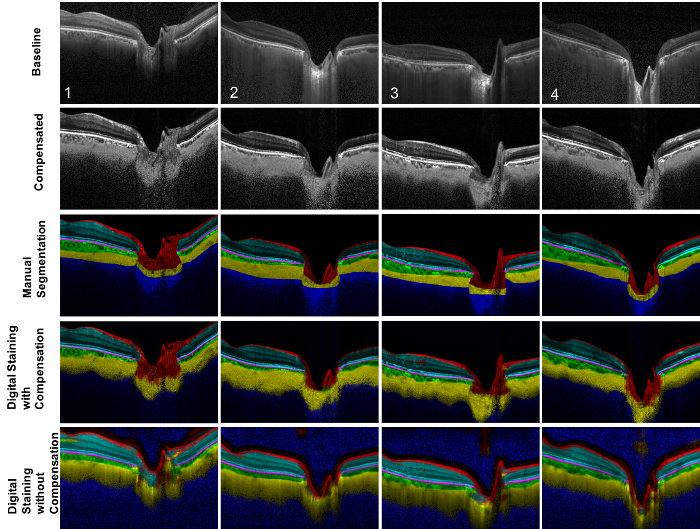}
    \caption{Baseline (1st row), compensated (2nd row), manually segmented (3rd row), digitally-stained images (trained on 10 compensated images; 4th row), and digitally-stained images (trained on 10 baselined images; 5th row) for 4 selected subjects (1: Healthy, 2$\&$3: POAG, 4:PACG ).}
    \label{fig:3}
\end{figure}

Baseline, compensated, manually segmented, and digitally-stained images (with training performed on 10 compensated or 10 baseline images) for 4 selected subjects (1: Healthy, 2$\&$3: POAG, 4: PACG ) can be found in Figure \ref{fig:3}.

When compensated images were used for training (Figure \ref{fig:3}, $4$th Row), we found that our digital staining algorithm was able to simultaneously highlight the RNFL + prelamina (in red), the RPE (in pink), all other retinal layers (in cyan), the choroid (in green), the sclera + LC (in yellow), and noise (in blue). Digitally-stained images were similar to those obtained from manual segmentation, and the results were consistent across all subjects and for all testing sets. Overall, the anterior LC was well captured, but our algorithm had a tendency to always identify LC insertions into the sclera that were not always present in the manual segmentations (e.g. subject 4). Small errors were sometimes observed. For instance, a small portion of the central retinal trunk was identified as choroidal tissue (green) in subject 2. Interestingly, while we provided a ``smooth" delineation of the choroid-scleral interface, our algorithm had a tendency to follow the ``undulations" of choroidal vessels. \\

When baseline images were instead used for training (Figure \ref{fig:3}, $5$th row), more errors were observed. For instance, parts of the retina and prelamina were identified as scleral tissue (yellow) in subject 1.

\subsection{Quantitative Analysis}
\begin{figure}[h]
    \centering
    \includegraphics[width=1.\textwidth]{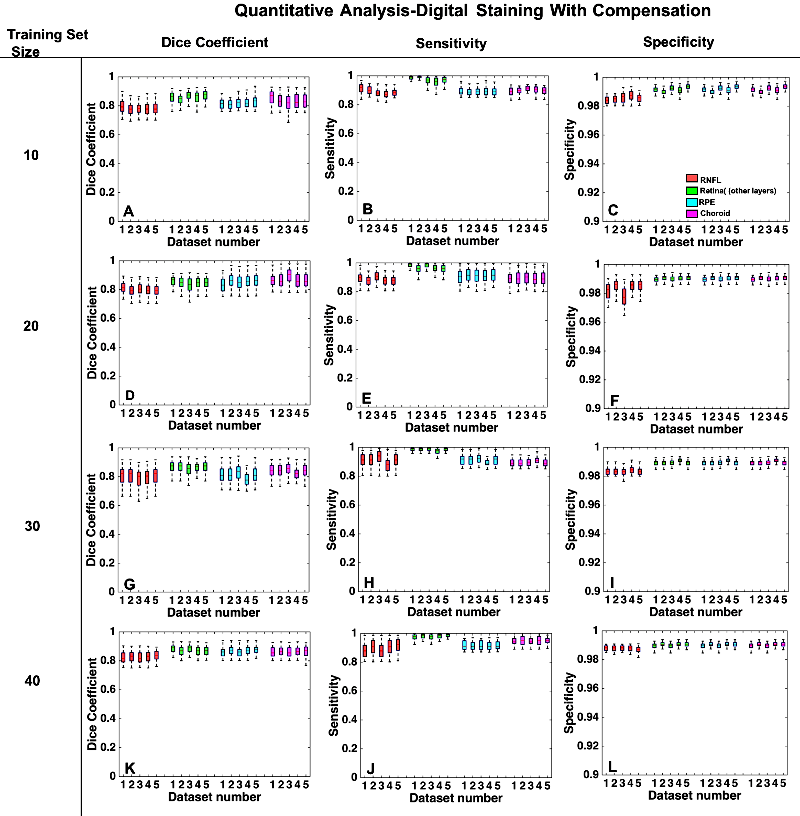}
    \caption{(A-C) Dice coefficients, sensitivities and specificities represented as box plots for each tissue (RNFL in red, all other retina layers in green, RPE in blue, and the choroid in magenta) for each of 5 testing set per tissue (training images per set: 10; testing images per set: 90). (D-F) Same as above except that 20 images were used for each training set, and 80 images for each testing set. (G-I) 30 images were used for each training set, and 70 images for each testing set. (K-L) 40 images were used for each training set, and 60 images for each testing set. }
    \label{fig:4}
\end{figure}

Across all tests (with training performed on compensated images), we found that the average dice coefficient was $0.82 \pm 0.05$ for the RNFL + prelamina, $0.84 \pm 0.02$ for the RPE, $0.86 \pm 0.03$ for all other retina layers, and $0.85 \pm 0.02$ for the choroid. Sensitivity and specificity were high for all tissues: $0.89 \pm 0.04$ and $0.99 \pm 0.00$ for the RNFL + prelamina, $0.90 \pm 0.03$ and $0.99 \pm 0.00$ for the RPE, $0.98 \pm 0.02$ and $0.99 \pm 0.00$ for all other retina layers, and $0.91 \pm 0.02$ and $0.99 \pm 0.00$ for the choroid respectively. For a given training set size, results were highly consistent across all tests (see Figure \ref{fig:4} showing Dice coefficients, sensitivities, and specificities for each tissue, 5 tests per tissue, training set size: 10). \\

For all tissues but RPE, we found that the training set size had no significant impact on the Dice coefficient and on sensitivity ($p > 0.05$ for all cases). However, increasing the training set size from 10 to 40 significantly improved the Dice coefficient for RPE ($p < 0.001$) from $0.81 \pm 0.04$ to $0.87 \pm 0.03$. Finally, the training set size had a significant impact on specificity for all tissues ($p < 0.001$ for all cases), however, we noted that specificity values were always higher than $0.98$ for all cases.

\begin{figure}[h]
    \centering
    \includegraphics[width=1.\textwidth]{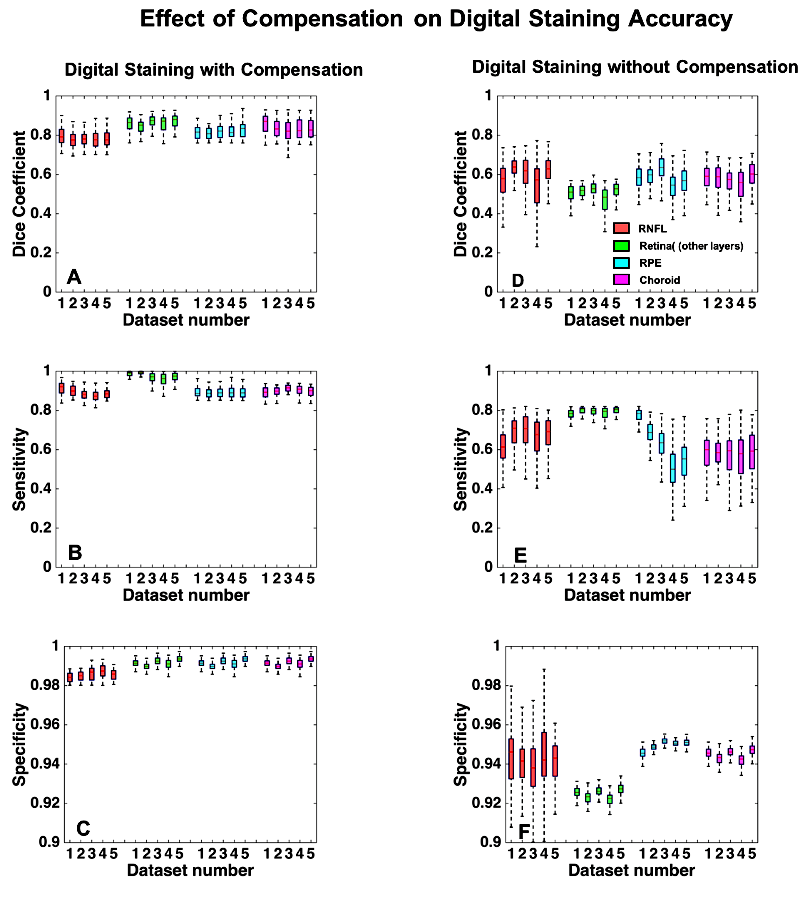}
    \caption{(A-C) Dice coefficients, sensitivities and specificities when our algorithm was trained on compensated images (10 images per training set). (D-F) Dice coefficients, sensitivities and specificities when our algorithm was trained on baseline (uncompensated) images (10 images per training set). 
}
    \label{fig:5}
\end{figure}

Overall, we found that digital staining performed significantly better when compensated images where used for training (as opposed to baseline or non-compensated images). Specifically, Dice coefficients, sensitivities, and specificities were always significantly higher when our algorithm was trained with compensated images (vs baseline images; $p < 0.001$ for all cases; Figure \ref{fig:5}).

\section{Discussion}

In this study, we have developed a custom deep learning algorithm to digitally stain tissues in OCT images of the ONH. Our algorithm was tested and validated using OCT images from 100 subjects, and was found to exhibit relatively good performance for all tissues. We also found that our algorithm performed significantly better when it was trained with enhanced (i.e. compensated) OCT images. Our method is attractive because it identifies all tissue layers simultaneously, and because it is ``universal", in the sense that the exact same processing approach is being used to identify each tissue layer individually. Our work could have applications for the clinical management of glaucoma using OCT. To the best of our knowledge, no deep learning techniques have yet been proposed for OCT images of the ONH.\\

In this study, we found that our digital staining algorithm was able to simultaneously isolate the RNFL + prelamina, the RPE, all other retina layers, the choroid, and the peripapillary sclera + LC. Our results were consistent across all subjects and we obtained relatively good agreements with respect to manual segmentations (for all tissues, averaged dice coefficients varied between 0.82 and 0.86, averaged sensitivities between 0.89 and 091, and averaged specificities were always higher than 0.98). We believe our work offers a framework to automatically measure structural parameters of the ONH. Typically, ONH tissues exhibit complex 3D structural changes during the development and progression of glaucoma, including, but not limited to: changes in RNFL thickness and minimum-rim-width \cite{RN35}, changes in LC depth,5 changes in LC curvature \cite{RN36}, changes in LC global shape index \cite{RN36}, changes in choroidal thickness \cite{RN2,RN3}, peripapillary atrophy \cite{RN37,RN38},  scleral canal expansion \cite{RN6}, migration of the LC insertion sites \cite{RN39,RN40},  LC focal defects \cite{RN41,RN42}, and scleral bowing \cite{RN7}. Obtaining these parameters from our digitally stained images should represent no major hurdles. Our approach may thus be of high importance for glaucoma diagnosis, management, and risk-profiling. \\

Although there exists accurate automated segmentation tools for retinal layers \cite{RN43,RN44} for the choroid \cite{RN12,RN13,RN18,RN45,RN46,RN47} or the choroid-scleral interface \cite{RN47}, and to a lesser extent, for the LC \cite{RN48,RN49,RN50}, there exists no ``universal" tool to isolate both connective and neural tissues simultaneously. Each tissue currently requires its own specific algorithm, each of which may be computationally expensive \cite{RN46}. Our current digital staining approach highlights all tissues simultaneously with the exact same deep learning backbone, and only requires a few seconds of processing time for each image on a standard GPU card. Note that our group is currently developing a real-time digital staining solution to make it more attractive for glaucoma clinics.\\

We found that the quality of digital staining was relatively poor when our deep learning network was trained with baseline (uncompensated) images (Figure \ref{fig:5}). On average (all tissues) the dice coefficient was $0.56 \pm 0.06$ (vs $0.84 \pm 0.03$ when training with compensated images), sensitivity $0.65 \pm 0.26$ (vs $0.92 \pm 0.03$), and specificity $0.93 \pm 0.02$ (vs $0.99 \pm 0.00$). This is not surprising as baseline images (vs compensated images) typically exhibits lower intra- and inter-layer contrasts, low visibility at high depth, and strong blood vessel shadow artefacts \cite{RN26}. Our work illustrates that adaptive compensation \cite{RN26} may be a necessary first step toward a simple solution to automatically segment the ONH tissues.\\

Interestingly, we found that increasing the size of our training set (from 10 to 40 images) did not significantly improve digital staining accuracy, except for the RPE. This result may appear counter-intuitive. However, contrarily to most deep learning applications, our situation is intrinsically low dimensional: most OCT scans of the ONH are fundamentally similar to each other (e.g. the sclera is always posterior to the choroid). This may potentially explain why our results were accurate even when training was performed on a small number of subjects (10). This can be seen as a strong advantage, as getting access to images is one the biggest limiting factor for deep learning applications. However, we believe that more validations are required, and we aim to study much larger populations in the near future. \\

Several limitations in our study warrant further discussion. First our algorithm was trained with OCT images from a single device (Spectralis), and it is currently unknown if our approach could be directly applied to images captured with other OCT devices. However, one could consider re-training the network for each device separately. We are currently exploring such an approach. \\

Second, the accuracy of our algorithms was assessed against manual segmentations from a single expert observer. Further work will be required to perform inter- and intra-observer variability tests and for multiple observers. Nevertheless, we offer here a proof of principle of ONH digital staining that could also be used by other groups for further validations.\\

Third, we were unable to provide an additional validation of our algorithm by comparing our stained images to those obtained from histology. This is extremely difficult to achieve as one would need to image a human ONH with OCT, process it with histology, and register both datasets. Note that the broad understanding of OCT ONH anatomy to histology has been based on a single comparison with a normal monkey eye scanned in vivo at an IOP of 10 mmHg and then perfusion fixed at time of sacrifice at the same IOP \cite{RN51}. The tissue classification derived from our algorithm matches the expected relationships observed in this canonical work. At the time of writing, there have been no published experiments matching human ONH histology to OCT images. While the absence of this work prevents an absolute validation of our technique, the same shortcoming necessarily applies to every other in vivo investigation of deep OCT imaging of the human investigation, many publications of which predate even the publication of the comparison with the monkey ONH.\\

Fourth, in some subjects, we observed false predictions for a few pixels in the LC. This shortcoming could potentially be addressed with the use of: (1) a deeper network; (2) a more advanced neural network architecture; (3) a 3D CNN; or (4) a simple post-processing approach to filter tissue discontinuities following the digital stain step. Further work is required to explore all these options.\\

Fifth, in the patch-based approach, overlapping patches result in multiple convolutions on similar sets of pixels which are a waste of computational memory and time. Recently developed architectures \cite{RN52,RN53}, for other biomedical imaging applications have circumvented these issues, which could be explored for OCT images of the ONH.\\

Sixth, in some subjects, the LC insertions into the sclera highlighted by the algorithm which were not visible to the expert observer during manual segmentations makes it unclear if our algorithm was able to identify faint signals that resembled LC insertions, or whether it was introducing artifacts. A 3D validation may be required to address this phenomenon.\\
 
In conclusion, we have developed a custom deep learning algorithm to digitally stain nervous and connective tissues in OCT images of the ONH. Because these tissues exhibit significant structural changes in glaucoma, digital staining may be of interest in the clinical management of glaucoma.

\section*{Acknowledgment}
Supported by the Singapore Ministry of Education Academic Research Funds Tier 1 (R-155-000-168-112; AT), a National University of Singapore (NUS) Young Investigator Award Grant (NUSYIA-FY13-P03; R-397-000-174-133; MJAG), and by the National Medical Research Council (Grant NMRC/STAR/0023/2014; TA).

\bibliographystyle{unsrt}
\bibliography{references}

\end{document}